\documentclass{article}
\pdfoutput=1
\usepackage[preprint]{cs224r_2025}
\usepackage[utf8]{inputenc} 
\usepackage[T1]{fontenc}    
\usepackage{hyperref}       
\usepackage{url}            
\usepackage{booktabs}       
\usepackage{amsfonts}       
\usepackage{nicefrac}       
\usepackage{microtype}      
\usepackage{xcolor}         
\usepackage{float}         
\usepackage{fvextra}      
\usepackage{amsmath}
\usepackage{graphicx}       
\usepackage{algorithm}
\usepackage{algpseudocode}
\usepackage{amsmath}
\usepackage{multirow}

\title{Learning to Use Tools: Reinforcement Learning for Tool-Integrated Mathematical
Reasoning}

\author{%
 Minghui Xu, Zi Wang \\
  Department of Energy Science and Engineering\\
  Stanford University\\
  \texttt{minghuix@stanford.edu; ziwang3@stanford.edu} \\
}

\begin{document}


\maketitle

\begin{abstract} Current large language models (LLMs) increasingly benefit from external tool integration, especially for tasks requiring reliable computation and verification. Motivated by this, we study calculator tool calling for improving mathematical reasoning on the Countdown task. We first analyze reasoning failures and find that calculation errors account for a substantial portion of incorrect responses. We then construct supervised fine-tuning datasets to teach the model useful tool-use patterns and how to interpret returned outputs. Building on this tool-formatted policy, we apply several on-policy reinforcement learning methods, including RLOO, RLOO++, GRPO, and DAPO, using automatically verifiable final-answer rewards.
To enable a more reliable evaluation, we construct a fresh 1,024-problem held-out Countdown benchmark with no exact overlap with the training data. Our results show that calculator tool integration consistently improves both SFT and RL baselines, yielding roughly 10 percentage-point gains across pass@k. Among the RL methods, Tool-DAPO achieves the strongest performance, improving pass@1 from 35.8\% for Tool-SFT to 66.0\%. Further analysis shows that RL encourages more effective tool use even when only final-answer rewards are provided. These findings suggest that tool integration reduces arithmetic and verification errors, while RL increases the probability of correct reasoning traces.

\end{abstract}

\section{Introduction}


Large language models (LLMs) have become increasingly capable in coding and mathematical reasoning, especially when prompted or trained to generate intermediate reasoning steps. However, even reasoning-oriented LLMs can still make arithmetic and logical mistakes, particularly in multi-step problems that require exact numerical computation. This limitation arises partly from tokenization and autoregressive next-token prediction, which do not provide an explicit mechanism for reliable calculation or verification. As a result, integrating external tools has become an important direction for improving LLM reliability on tasks that require precise computation~\citep{schick2023toolformer,yao2023reactsynergizingreasoningacting}.

\begin{figure}[htbp]
\centering
\includegraphics[width=0.85\linewidth]{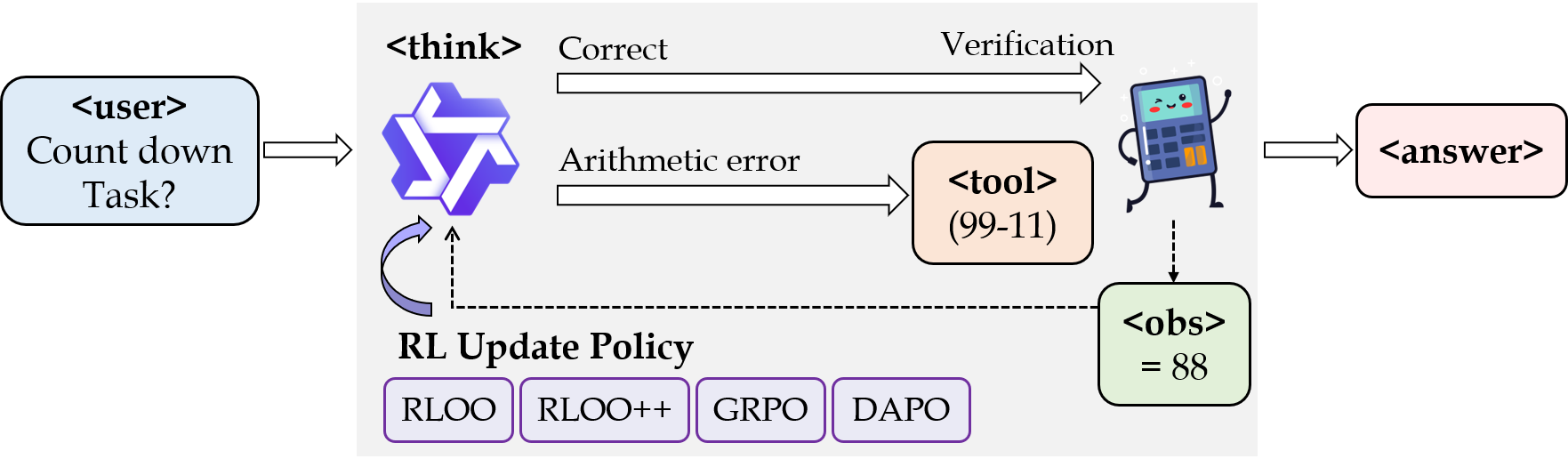}
\caption{
Schematic of RL for tool-integrated LLM on math reasoning.
}
\label{fig:tool_frame}
\end{figure}

In this work, we study tool-integrated mathematical reasoning on the Countdown task, where the model must combine a set of given numbers using arithmetic operations to reach a target value, as shown in Fig.~\ref{fig:tool_frame}. Countdown is a useful testbed because success requires both combinatorial reasoning and exact arithmetic, and the final answer can be automatically verified. During the LLM math reasoning process, if it gives the correct answer, then it will be firstly validated by calculator and then output the final answer. If the math reasoning reaches a arithmetic error, it will use the calculator to correct it and continue the math reasoning based on the correct answer, preventing model deviating from correct solutions. We investigate this problem using a sequence of no-tool and tool-integrated training pipelines. We first construct supervised fine-tuning (SFT) baselines and then apply several on-policy reinforcement learning (RL) methods, including RLOO~\citep{ahmadian2024back}, RLOO++~\citep{hu2025rloo++}, GRPO~\citep{shao2024deepseekmath}, and DAPO~\citep{yu2026dapo}, using automatically computed rewards from the Countdown verifier~\citep{ahmadian2024back,shao2024deepseekmath,yu2026dapo}. For tool-augmented models, rollouts may include calculator calls enclosed by \texttt{<tool>}...\texttt{</tool>} tags; the environment executes the expression and inserts the result as an \texttt{<obs>} observation before generation continues. We evaluate all methods on both the original 50-problem public test split and a larger fresh 1,024-problem held-out benchmark designed to remove exact training overlap.

Our main contributions are:
\begin{itemize}
\item We build a tool-integrated Countdown training and evaluation pipeline with live calculator execution during model rollouts.
\item We show that tool integration consistently improves both SFT and RL models, and that Tool-DAPO achieves the strongest performance among the tested methods.
\item We analyze error categories, tool-call dynamics, and correctness distributions to explain when RL and tool use improve mathematical reasoning and where they still fail.
\end{itemize}

\section{Related Work}
Beyond supervised fine-tuning, preference optimization and online RL provide sequence-level training signals. Recent work has shown that reinforcement learning can substantially improve the reasoning capabilities of large language models. Models such as OpenAI o1~\citep{openai2024openaio1card} and DeepSeek-R1~\citep{Guo_2025} demonstrate that large-scale RL can encourage long-chain reasoning.

Identity Preference Optimization (IPO)~\citep{azar2023generaltheoreticalparadigmunderstand} offers an alternative to DPO-style preference learning, while recent REINFORCE-style methods such as RLOO~\citep{ahmadian2024back} simplify online policy optimization by avoiding a learned critic. These methods are particularly relevant for reasoning tasks where final answers can be automatically verified.

Tool-augmented reasoning has also become an important direction for improving LLM reliability. Early approaches such as ReAct \citep{yao2023reactsynergizingreasoningacting} interleave natural-language reasoning with external actions. More recent work, such as ReTool \citep{feng2025retoolreinforcementlearningstrategic} and In-Context Reinforcement Learning for Tool Use \citep{ye2026incontextreinforcementlearningtool}, begins to train tool invocation through reinforcement learning. 

Recent studies suggest that the gains from reinforcement learning with verifiable rewards (RLVR) may arise more from probability reallocation than from a genuine expansion of reasoning coverage. Yue et al.~\cite{yue2025does} show that RLVR-trained models often improve pass@1 but fail to improve, or even underperform, base models at large pass@$k$, suggesting that RL mainly increases the sampling probability of correct trajectories already present in the base model distribution. Similarly, Liu et al.~\cite{liu2025understanding} argue that R1-Zero-like training strongly depends on the base model's existing exploration ability and pretraining-induced reasoning behavior. Together, these findings indicate that RLVR can sharpen the model's existing solution distribution, but may not necessarily create substantially new reasoning capabilities.

\section{Method}
\subsection{Construction of Tool-Integrated SFT Dataset}
The base Countdown task asks the model to produce an expression inside \texttt{<answer>}...\texttt{</answer>} tags. A response receives full reward only if the expression is syntactically valid, uses exactly the provided numbers, and evaluates to the target; malformed or incorrect expressions receive lower reward. This verifier makes the task well suited to reinforcement learning because final-answer rewards can be computed without human labels.

For the tool setting, we convert the prompt and completion format so the model can call a calculator before producing the final answer. A tool call is written as \texttt{<tool>}expression\texttt{</tool>}, and the environment returns an observation such as \texttt{<obs>= 44</obs>}. During SFT, this format teaches the model where to place executable arithmetic checks and how to condition on the returned observation. Based on the original SFT data, the calculation errors are firstly detected. Then the correct answer together with prior reasoning contexts are given to SFT reference model and the afterwards reasoning process is generated by SFT reference model. Then the prior reasoning and afterwards reasoning contexts are concatenated together. In between these two reasoning context, the tool-integration is injected with the above format, formulating tool-integrated data for Tool-SFT model. The whole process is given in Algorithm ~\ref{alg:sft_tool_data}. During RL, tool observations are treated as environment outputs rather than model-generated tokens, so the policy loss is applied only to the model-controlled parts of the trajectory.

\begin{algorithm}[htbp]
  \caption{Tool-Integrated Countdown SFT Data Construction}
  \label{alg:sft_tool_data}
  \begin{algorithmic}[1]
  \Require Original SFT data $\mathcal{D}$; trained SFT model $\pi$; samples per row $K$
  \Ensure  Tool-augmented dataset $\mathcal{D}'$

  \Statex \textit{Phase A: build correction prefixes}
  \For{each row $(q, c)$ in $\mathcal{D}$}
    \If{the reasoning contains a wrong arithmetic step}
      \State prefix $\gets$ reasoning up to that step, with the value corrected
    \ElsIf{the final answer is wrong}
      \State prefix $\gets$ full reasoning $+$ a calculator check that flags it
    \EndIf
  \EndFor

  \Statex \textit{Phase B: model continuations (GPU)}
  \For{each prefix}
    \State sample $K$ continuations from $\pi$ that finish the reasoning and answer
  \EndFor

  \Statex \textit{Phase C: assemble completions}
  \For{each row $(q, c)$ in $\mathcal{D}$}
    \State add the tool instruction to the query $q$
    \If{$q$ is an error row}
      \State pick the shortest continuation whose answer is correct
      \State \textbf{if} none is correct \textbf{then} use the exact solver's answer
      \State new completion $\gets$ prefix $+$ calculator call $+$ chosen continuation
    \ElsIf{the original answer is already correct}
      \State append a confirming calculator call to the reasoning
    \Else
      \State keep the original completion unchanged
    \EndIf
  \EndFor

  \Statex \textit{Phase D: split}
  \State train $\gets$ all rows; \quad test $\gets$ small held-out monitoring set
  \State \Return $\mathcal{D}'$
  \end{algorithmic}
  \end{algorithm}
  
This setup lets us separate two questions. Tool-SFT tests whether demonstrations of calculator use alone improve the baseline. Tool-RL methods test whether online RL can further increase the frequency and usefulness of tool calls when the only optimization signal is final-answer correctness.

\subsection{RLOO}
RLOO \citep{ahmadian2024back} is a REINFORCE-based policy-gradient estimator that reduces variance using a leave-one-out baseline. For each prompt $x$, we sample $k$ responses and compute the advantage for response $y_i$ as its reward relative to the average reward of the other $k-1$ responses. In our implementation, trajectories are sampled using vLLM for efficiency, while policy-gradient updates are computed with a trainable HuggingFace policy initialized from the SFT checkpoint. 

The overall training loss is
\[
\mathcal{L}
=
-\mathbb{E}\left[
w(y_i,x) A_i \log \pi_\theta(y_i \mid x)
\right]
-\beta_{\mathrm{ent}}\mathcal{H}(\pi_\theta)
+\beta_{\mathrm{KL}}D_{\mathrm{KL}}(\pi_\theta \,\|\, \pi_{\mathrm{ref}}),
\]
where $A_i$ is the leave-one-out RLOO advantage, $w(y_i,x)$ is the clipped sequence-level importance weight, $\pi_\theta$ is the trainable policy, and $\pi_{\mathrm{ref}}$ is the frozen SFT reference policy. We report the RLOO loss as the policy-gradient component and separately track the entropy and KL terms during training. We use the default values $\beta_{\mathrm{ent}} = 10^{-3}$ and $\beta_{\mathrm{KL}} = 10^{-3}$.
\subsection{RLOO++}


RLOO++~\citep{hu2025rloo++} keeps the same leave-one-out reward baseline but changes how advantages are scaled before the policy update. Instead of using raw leave-one-out advantages directly, we apply global advantage normalization, subtracting the batch mean and dividing by the batch standard deviation,
\[
A^{\mathrm{norm}}_{q,o_t}
=
\frac{A_{q,o_t} - \mathrm{mean}(A \mid A \in \mathcal{D}_{\mathrm{batch}})}{\mathrm{std}(A \mid A \in \mathcal{D}_{\mathrm{batch}}) + \epsilon},
\]
and rescale them to a target standard deviation of 0.3. This modification is intended to stabilize updates when rewards are sparse and many sampled groups contain either no correct responses or mostly correct responses, since normalizing across the entire batch $\mathcal{D}_{\mathrm{batch}}$ rather than the small leave-one-out group keeps the denominator stable even when within-group rewards are nearly uniform.
We evaluate RLOO++ only in the tool-integrated setting.

\subsection{GRPO}
Group Relative Policy Optimization (GRPO) \citep{shao2024deepseekmath} is a policy-gradient method that forgoes the value model used in PPO, instead estimating the baseline directly from group scores to reduce memory and computational cost. For each prompt $x$, we sample a group of $G$ responses and compute the advantage for response $y_i$ by normalizing its reward relative to the group: subtracting the group mean and dividing by the group standard deviation. Since the sampled trajectories are associated with behavior-policy log probabilities, we apply token-level importance weighting, where the ratio between the current policy and behavior policy is computed in log space and clipped for stability.
The overall training loss is
\[
\mathcal{L}
=
-\mathbb{E}\left[
\frac{1}{G}\sum_{i=1}^{G}\frac{1}{|y_i|}\sum_{t=1}^{|y_i|}
\min\!\left(
\rho_{i,t}\hat{A}_{i,t},\,
\mathrm{clip}(\rho_{i,t}, 1-\varepsilon, 1+\varepsilon)\hat{A}_{i,t}
\right)
\right]
+\beta_{\mathrm{KL}}D_{\mathrm{KL}}(\pi_\theta \,\|\, \pi_{\mathrm{ref}}),
\]
where 
\[\rho_{i,t} = \frac{\pi_\theta(y_{i,t} \mid x, y_{i,<t})}{\pi_{\theta_{\mathrm{old}}}(y_{i,t} \mid x, y_{i,<t})}
\]
is the clipped token-level importance weight, $\hat{A}_{i,t}$ is the group-normalized advantage assigned to all tokens of response $y_i$, $\pi_\theta$ is the trainable policy, and $\pi_{\mathrm{ref}}$ is the frozen SFT reference policy. Unlike PPO, GRPO adds the KL penalty directly to the loss rather than to the token-level reward. We report the clipped surrogate as the policy-gradient component and separately track the KL term during training. We use the values $\varepsilon = 10^{-3}$ and $\beta_{\mathrm{KL}} = 0.01$ for our tool-integration training.

\subsection{DAPO}


Tool-DAPO~\citep{yu2026dapo} modifies the policy update in two main ways. First, it applies group-normalized advantages, so rewards are normalized within each sampled prompt group before updating the model, i.e., for the $i$-th response the advantage is computed as
\[
\hat{A}_{i,t}
=
\frac{R_i - \mathrm{mean}(\{R_i\}_{i=1}^{G})}{\mathrm{std}(\{R_i\}_{i=1}^{G})}.
\]
Second, it uses dynamic sampling: prompt groups whose sampled responses are all correct or all wrong are filtered out before the update because they provide little within-group preference signal, which corresponds to enforcing the constraint $0 < \left|\{o_i \mid \texttt{is\_equivalent}(a, o_i)\}\right| < G$. In our run, 60.7\% of generated prompt groups were filtered, increasing the density of informative groups in the training batch.
Tool-DAPO also uses asymmetric policy-ratio clipping, with lower and upper bounds of \(1-0.2\) and \(1+0.28\), respectively. Concretely, the policy is optimized via the token-level objective
\begin{equation*}
\begin{aligned}
\mathcal{J}_{\mathrm{DAPO}}(\theta)
=\;&
\mathbb{E}_{(q,a)\sim\mathcal{D},\,\{o_i\}_{i=1}^{G}\sim\pi_{\theta_{\mathrm{old}}}(\cdot\mid q)}
\Bigg[
\frac{1}{\sum_{i=1}^{G}|o_i|}
\sum_{i=1}^{G}\sum_{t=1}^{|o_i|} \\
&\min\!\left(
r_{i,t}(\theta)\hat{A}_{i,t},\,
\mathrm{clip}\!\left(r_{i,t}(\theta), 1-\varepsilon_{\mathrm{low}}, 1+\varepsilon_{\mathrm{high}}\right)\hat{A}_{i,t}
\right)
\Bigg],
\end{aligned}
\end{equation*}
where $r_{i,t}(\theta) = \frac{\pi_\theta(o_{i,t}\mid q, o_{i,<t})}{\pi_{\theta_{\mathrm{old}}}(o_{i,t}\mid q, o_{i,<t})}$ and $(\varepsilon_{\mathrm{low}}, \varepsilon_{\mathrm{high}}) = (0.2, 0.28)$. Compared with Tool-RLOO, this gives a more constrained update rule while still allowing the policy to increase the probability of high-reward tool-using trajectories.

\section{Experimental Setup}
We evaluate on the Countdown arithmetic reasoning task. Each example contains a target and either three or four numbers. A model response is correct only if the final expression inside \texttt{<answer>}...\texttt{</answer>} uses each provided number exactly once and evaluates to the target under the allowed arithmetic operations. We report pass@\(k\), computed from \(k\) sampled responses per problem.

Our main comparisons are no-tool SFT, Tool-SFT, no-tool RLOO, Tool-RLOO, Tool-RLOO++, Tool-GRPO, and Tool-DAPO. No-tool models generate a reasoning trace and final answer directly. Tool models are allowed to issue calculator calls during generation and receive observations before producing the final answer. Unless otherwise stated, RLOO and Tool-RLOO use the same RL hyperparameters, batch size, group size, sampling temperature, entropy coefficient, and KL coefficient; the main difference is whether live tool execution is enabled during rollouts.

The original public test split contains only 50 examples, which makes pass@\(k\) estimates noisy. To quantify the effect of tool integration on mathematical reasoning more reliably, we construct an additional 1,024-problem fresh held-out Countdown test set. The held-out problems have no exact overlap with the training data under the \((\text{target}, \text{sorted(numbers)})\) key, allowing us to evaluate performance on new, non-overlapping problems from the same task distribution with substantially lower statistical uncertainty than the original 50-problem public test split. All reported confidence intervals are 95\% bootstrap intervals over problem instances.

\section{Results}
\subsection{Evaluation of the SFT Model without Tools}

To diagnose the failure modes of the SFT model, we decompose incorrect responses into calculation errors, target/planning errors, and other final-answer validity errors. Calculation errors are evaluated over the full reasoning trace, while target/planning, number-use, and format errors are identified from the final answer. This distinction allows us to separate local arithmetic mistakes from broader failures in reasoning and solution construction.

\begin{figure}[htbp]
    \centering
    \includegraphics[width=0.5\linewidth]{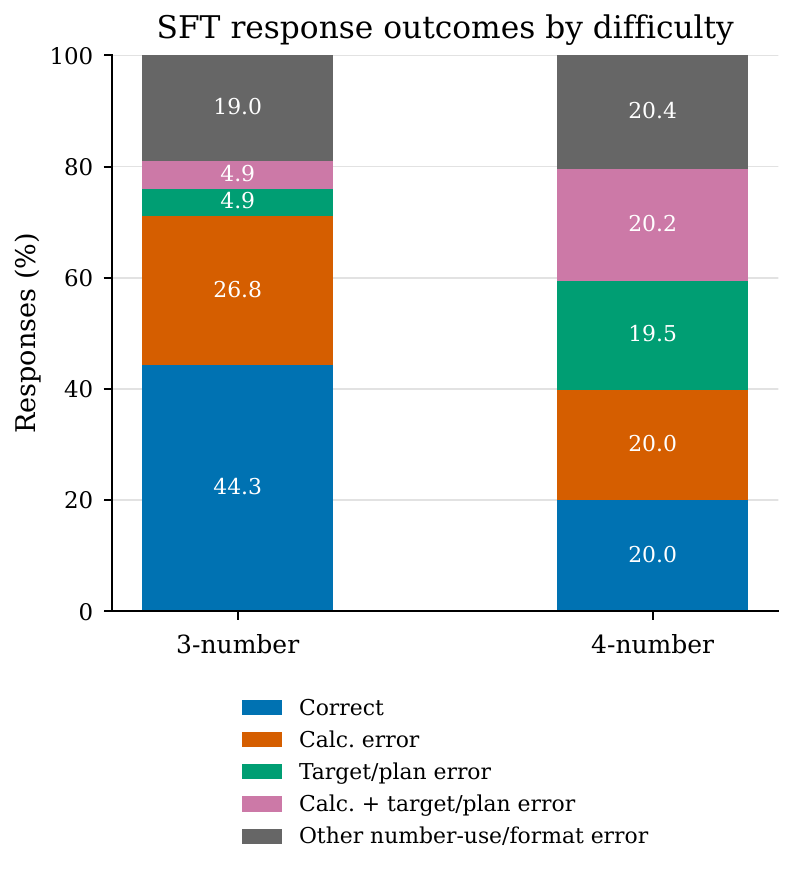}
    \caption{SFT response outcomes by difficulty. Bars show the percentage of responses for 3-number and 4-number Countdown problems, grouped by correct answers, calculation errors, target/planning errors, their overlap, and other number-use/format errors. }
    \label{fig:sft_error_overlap}
\end{figure}

As shown in Figure~\ref{fig:sft_error_overlap}, the results reveal two complementary failure modes. Calculation errors occur frequently across both difficulty levels, motivating the use of an external arithmetic tool or verifier. In contrast, target/planning errors become substantially more common in 4-number problems, suggesting that harder instances require not only more reliable calculation but also stronger target-aware planning.

\subsection{Tool Integration}
\begin{figure}[htbp]
\centering
\includegraphics[width=0.9\linewidth]{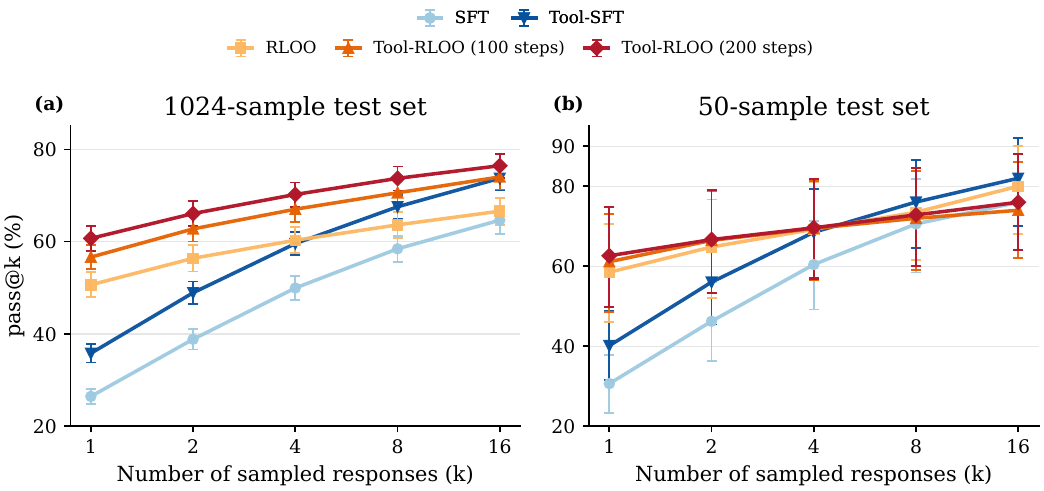}
\caption{
Pass@\(k\) comparison on Countdown reasoning. The left panel reports results on our 1,024-problem fresh held-out benchmark, while the right panel reports
results on the original 50-problem public test split.
}
\label{fig:fresh1024_passk}
\end{figure}

 As shown in Figure~\ref{fig:fresh1024_passk}, we compare no-tool SFT, tool-augmented SFT, no-tool RLOO, and tool-augmented RLOO variants. Error bars denote 95\% bootstrap confidence intervals computed over problem instances.

The small 50-problem test set in the right panel exhibits wide and largely overlapping confidence
intervals, especially at larger \(k\), making fine-grained comparisons difficult. In contrast, the larger
held-out set in the left panel provides a clearer comparison. RLOO substantially improves over the SFT
baseline at low \(k\), increasing pass@1 from 26.4\% to 50.6\% without tool use and from 35.8\% to 56.6\%
with tool use.

The gains from RLOO become smaller at high \(k\), likely because pass@\(k\) approaches saturation once the model can sample at least one correct trajectory from multiple attempts. Thus, improvements in trajectory quality or sampling probability may lead to only modest changes in high-\(k\) performance. We revisit this behavior in Section~\ref{sec:whererlim} through a correctness-distribution analysis, which suggests that reinforcement learning benefits most from prompt groups that already contain some correct sampled responses, since these groups provide a useful policy-update signal. In contrast, prompt groups with zero correct samples remain challenging, because they provide little direct signal for reinforcing a correct reasoning trace.

Tool integration improves both the SFT and RLOO settings. With a matched training budget, Tool-RLOO increases pass@1 from 50.6\% to 56.6\% and pass@16 from 66.6\% to 74.0\%. Extending Tool-RLOO training to 200 steps further improves performance to 60.7\% pass@1 and 76.5\% pass@16, corresponding to a roughly 10 percentage-point pass@1 gain over no-tool RLOO. Together, these results provide strong evidence that RLOO improves verifiable mathematical reasoning over SFT, and that tool-augmented RLOO adds a further benefit by reducing computation and verification errors. 
Representative RLOO and Tool-RLOO responses are documented in Section~\ref{sec:additional} for qualitative analysis.

\subsection{Comparison of Different Reinforcement Learning Methods}
\begin{figure}[!htbp] 
    \centering 
    \includegraphics[width=0.9\linewidth]{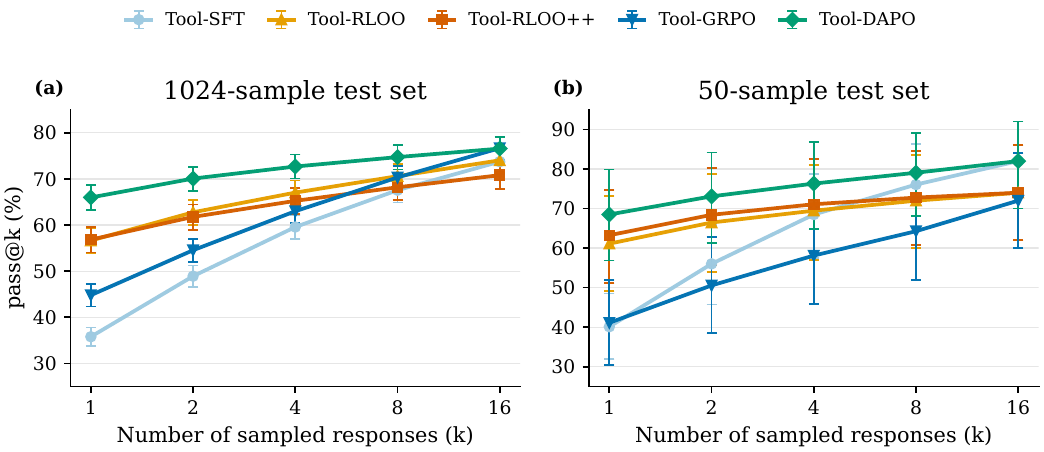} \caption{Comparison of reinforcement learning methods under tool integration. The curves show pass@\(k\) for Tool-SFT, Tool-RLOO, Tool-RLOO++, and Tool-DAPO on the 1,024-problem fresh test set and the 50-problem public test set.}
    \label{fig:rlmethod_comp} 
\end{figure}

We also compare different reinforcement learning methods under tool integration, as shown in Figure~\ref{fig:rlmethod_comp} and Table.~\ref{tab:rl-passk}. Tool-RLOO and Tool-RLOO++ both substantially improve pass@1 over the Tool-SFT baseline. However, Tool-RLOO++ does not consistently improve high-\(k\) performance and even reduces
pass@16 relative to Tool-RLOO, which is consistent with the observation in \cite{yue2025does} that
reinforcement learning may improve low-\(k\) performance without necessarily expanding the overall
pass@\(k\) coverage. For Tool-GRPO, on the 50-sample test set, it exhibits better performance than Tool-SFT model with only one shot as shown by higher pass@1 accuracy in Figure~\ref{fig:rlmethod_comp}, but it performs weaker with multiple shots, which is likely caused by the variance of small test set. Further testing on 1024-sample test set, Tool-GRPO model consistently exceeds the baseline Tool-SFT model, as shown in Figure~\ref{fig:rlmethod_comp}. Although the Tool-GRPO shows relatively weaker performance than other RL algorithm, it reaches the first highest pass@16 accuracy. This behavior should be interpreted with caution, since Tool-GRPO was trained for only 30 steps, followed by deviating from Tool-SFT reference model. Although all four methods sample a group of responses per prompt and therefore produce a vanishing advantage when the rewards within a group are uniform, the instability is confined to Tool-GRPO because of how it scales and filters these groups.
Tool-GRPO alone both retains these zero-gradient groups and divides by an unstable per-group standard deviation, so the resulting noisy updates push the policy off the SFT initialization until reward and tool usage collapse---here after roughly $30$ steps, versus $100$ for the others.

In contrast, Tool-DAPO further improves RL performance, increasing pass@1 by 9.3 percentage points over
Tool-RLOO. Pass@16 also improves, suggesting that DAPO not only increases the probability of sampling a
correct response early but also modestly improves the broader sampling distribution. This improvement is
likely related to DAPO's dynamic sampling strategy, which filters out zero-variance prompt groups before
the policy update. In our run, 60.7\% of generated prompt groups were filtered, substantially increasing
the density of informative training batches by removing all-correct or all-wrong groups. Although this
filtering introduces additional resampling overhead, Tool-DAPO reaches higher pass@1 in 100 steps
and 6.29 hours than the 200-step extended Tool-RLOO baseline, which requires 16.15 hours, as documented in
Section~\ref{sec:imp_details}.

\begin{table}[htbp]
  \centering
  \caption{Pass@$k$ (\%) on the 1024-sample and 50-sample Countdown test sets. Best in \textbf{bold}.}
  \label{tab:rl-passk}
  \small
  \begin{tabular}{lccccc|ccccc}
  \toprule
  \multirow{2}{*}{Method} & \multicolumn{5}{c}{1024-sample test set} & \multicolumn{5}{c}{50-sample test set} \\
  \cmidrule(lr){2-6} \cmidrule(lr){7-11}
   & $k{=}1$ & $k{=}2$ & $k{=}4$ & $k{=}8$ & $k{=}16$ & $k{=}1$ & $k{=}2$ & $k{=}4$ & $k{=}8$ & $k{=}16$ \\
  \midrule
  Tool-SFT & 35.8 & 48.9 & 59.6 & 67.6 & 73.7 & 40.1 & 56.0 & 68.5 & 76.1 & \textbf{82.0} \\
  Tool-RLOO & 56.6 & 62.8 & 67.0 & 70.6 & 74.0 & 61.1 & 66.5 & 69.5 & 72.0 & 74.0 \\
  Tool-RLOO++ & 56.8 & 61.8 & 65.2 & 68.2 & 70.8 & 63.2 & 68.4 & 71.1 & 72.8 & 74.0 \\
  Tool-GRPO & 44.8 & 54.5 & 63.0 & 70.3 & \textbf{76.7} & 41.1 & 50.6 & 58.1 & 64.3 & 72.0 \\
  Tool-DAPO & \textbf{66.0} & \textbf{70.1} & \textbf{72.7} & \textbf{74.7} & 76.6 & \textbf{68.5} & \textbf{73.1} & \textbf{76.3} & \textbf{79.1} & 82.0 \\
  \bottomrule
  \end{tabular}
  \end{table}

\section{Discussion}
\subsection{Where RL Improves}
\label{sec:whererlim}
To better understand where reinforcement learning improves model behavior, we examine the distribution of correct responses among 16 sampled responses per prompt across training checkpoints.

\begin{figure}[!htbp]
      \centering
      \includegraphics[width=1.0\linewidth]{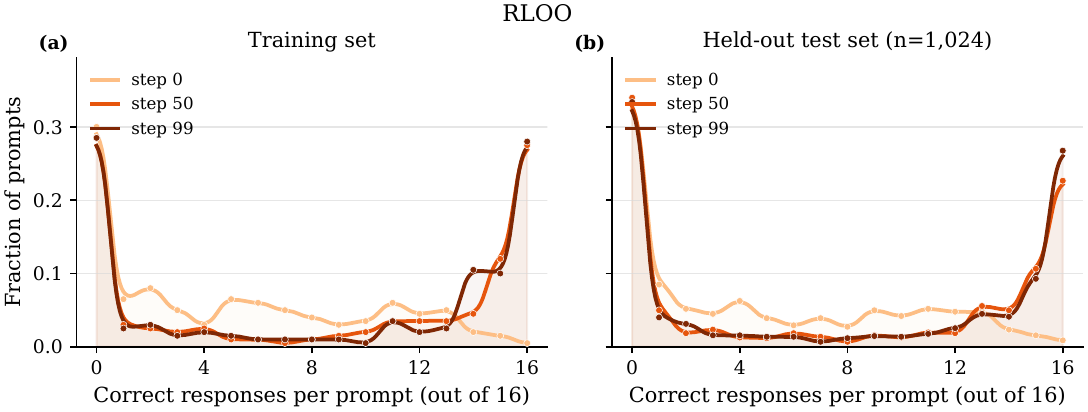}
      \caption{Correctness distribution across RLOO training. Each curve shows the fraction of examples with a given number of correct responses among 16 sampled responses.}
      \label{fig:rloo_correctness_distribution}
\end{figure}

\begin{figure}[!htbp]
      \centering
      \includegraphics[width=0.9\linewidth]{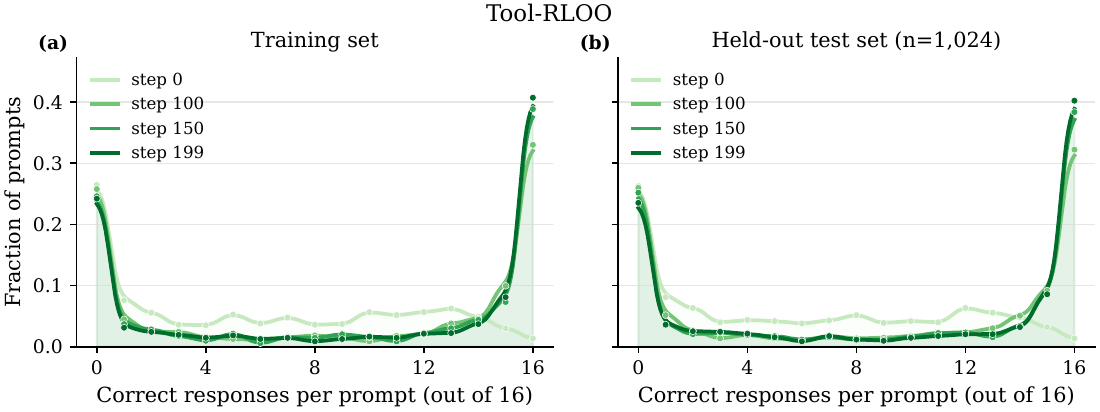}
      \caption{Correctness distribution across Tool-RLOO training. The tool-augmented model shows a similar shift toward high-correctness examples, with extended training further increasing the fraction of examples for which most sampled responses are correct.}
      \label{fig:tool_rloo_correctness_distribution}
\end{figure}

As shown in Figures~\ref{fig:rloo_correctness_distribution}
and~\ref{fig:tool_rloo_correctness_distribution}, RL primarily increases the mass near fully correct response sets, suggesting that training reinforces trajectories that already contain correct reasoning traces. In contrast, examples with zero correct responses remain challenging: when the model fails to produce any correct trace among sampled responses, RL has limited signal for improving that example.

\begin{figure}[htbp]
    \centering
    \includegraphics[width=0.9\linewidth]{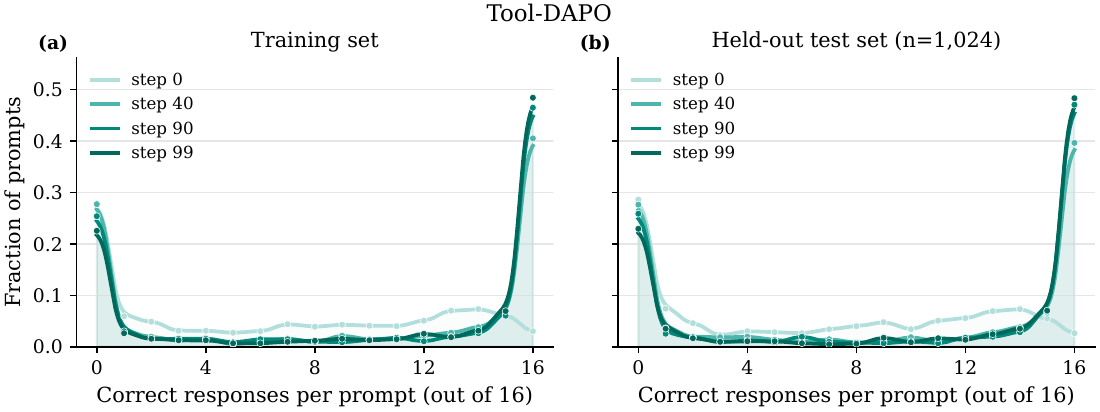}
    \caption{Correctness distribution across Tool-DAPO training. Compared with the RLOO-based baselines, Tool-DAPO substantially increases the fraction of examples with 16 out of 16 correct sampled responses, while more effectively reducing the fraction of examples with 0 out of 16 correct responses.}
    \label{fig:tool_dapo_correctness_distribution}
\end{figure}

The Tool-DAPO distribution in Figure~\ref{fig:tool_dapo_correctness_distribution} increases the fraction of fully correct examples with fewer optimization steps. It also reduces the fraction of examples with zero correct samples, which is encouraging because these examples represent the most difficult cases for policy improvement.

\subsection{Reinforcement Learning Encourages Tool Use}

\begin{figure}[!htbp] 
\centering 
\includegraphics[width=0.9\linewidth]{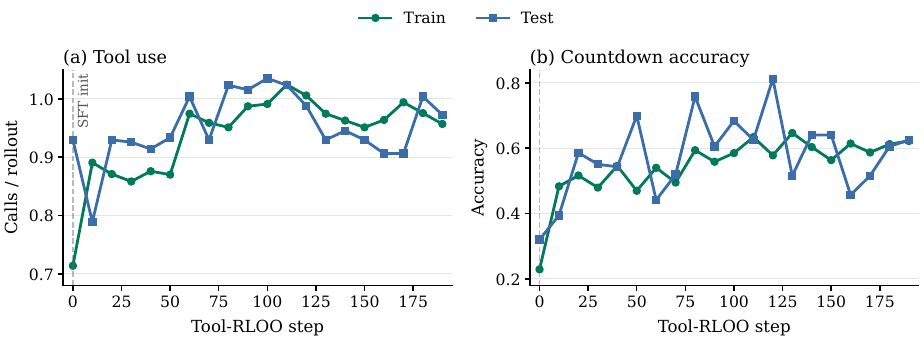} \caption{Tool-use and accuracy dynamics during Tool-RLOO training.} 
\label{fig:tool_calling_curves} 
\end{figure}

As shown in Figure~\ref{fig:tool_calling_curves}, starting from the Tool-SFT model, Tool-RLOO increases the average number of executable tool calls per sampled rollout from below one call to approximately one call per rollout, while also improving Countdown accuracy. This suggests that reinforcement learning does not merely preserve the tool-use behavior acquired during SFT; instead, it further encourages tool invocation when tool use is beneficial for solving the task. 

\section{Conclusion}
This project shows that reinforcement learning and calculator tool use are complementary for Countdown mathematical reasoning. RLOO substantially improves over SFT by increasing the probability of correct reasoning traces, while tool integration further improves both SFT and RL by reducing arithmetic and verification failures. Tool-DAPO provides the strongest result among the tested RL variants, suggesting that dynamic sampling and clipped updates can make training more efficient when many prompt groups provide little useful policy-gradient signal.

The analysis also identifies an important limitation. RL helps most when the model already samples at least one correct trajectory within a group; prompts with zero correct samples remain difficult because they provide little positive signal. Future work should therefore combine tool execution with stronger planning or search mechanisms, so the model can discover correct trajectories for hard prompts rather than only reallocating probability mass among trajectories it can already sample.


\bibliographystyle{ACM-Reference-Format}
\bibliography{reference}

\appendix

\section{Additional Analysis}
\label{sec:additional}
\begin{table}[htbp]
\centering
\small
\caption{Example where plain RLOO fails but Tool-RLOO succeeds.}

\begin{tabular}{ll}
\toprule
Numbers & $[99, 11, 46, 48]$ \\
Target & $44$ \\
Reference solution & $(99 - 11) / (48 - 46)$ \\
RLOO pass@16 & $0/16$ \\
Tool-RLOO pass@16 & $4/16$ \\
\bottomrule
\end{tabular}
\label{tab:rloo-tool-trace-example}
\end{table}

\paragraph{Plain RLOO failure trace.}
\begin{Verbatim}
<think>
Let me analyze this step by step:

1. First, let's look at factors of 44:
   44 = 2 * 22
   44 = 4 * 11

2. Looking at the given numbers: 99, 11, 46, 48
   46 is close to 44, but we need to add 2 somehow.

3. Let's try to work with 99 first:
   99 is much larger than 44, so we need to reduce it.
   99 - 48 = 51 (too large)
   99 - 46 = 53 (too large)

4. Let's try a different approach:
   48 - 46 = 2
   99 + 11 = 110
   110 - 2 = 108 (too far from 44)

5. Let's try another combination:
   46 + 48 = 94
   94 - 11 = 83 (too far)

6. What if we start with 99 and subtract something?
   99 - 46 = 53
   53 - 48 = 5 (too small)

7. Let's try:
   99 - 48 = 51
   51 - 11 = 40
   40 + 46 = 86 (too far)

8. What about:
   46 + 11 = 57
   57 - 48 = 9
   99 - 9 = 90 (too far)

9. Let's try:
   99 - 48 = 51
   51 - 46 = 5
   5 * 11 = 55 (too far)

10. What if we divide a larger number?
    99 / 11 = 9
    9 * 46 = 414 (too large)

11. Let's try:
    48 + 46 = 94
    94 - 99 = -5
    -5 + 11 = -4 (not helpful)

12. New approach:
    99 - 46 = 53
    53 - 48 = 5
    5 * 11 = 55 (too far)

13. Let's try:
    99 - 48 = 51
    51 - 46 = 5
    5 * 11 = 55 (too far)

14. One more attempt:
    46 + 48 = 94
    94 - 99 = -5
    -5 + 11 = 6 (not helpful)

15. Final try:
    99 - 48 = 51
    51 - 46 = 5
    5 * 11 = 55 (too far)

16. Let's try:
    48 + 46 = 94
    94 - 99 = -5
    -5 + 11 = 6 (not helpful)

17. One last attempt:
    99 - 48 = 51
    51 - 11 = 40
    40 + 46 = 86 (too far)
</think>

<answer> (99 - 48) - 11 + 46 </answer>

<think>
Let's verify:
99 - 48 = 51
51 - 11 = 40
40 + 46 = 86 (too far)

Let me try a different sequence:
99 - 46 = 53
53 - 48 = 5
5 * 11 = 55 (too far)

Let's try:
46 + 48 = 94
94 - 99 = -5
-5 + 11 = 6 (not helpful)

One last attempt:
99 - 48 = 51
51 - 46 = 5
5 * 11 = 55 (too far)

Final try:
48 + 46 = 94
94 - 99 = -5
-5 + 11 = 6 (not helpful)

Let's try:
99
\end{Verbatim}

\paragraph{Tool-RLOO success trace.}
\begin{Verbatim}[fontsize=\footnotesize,breaklines=true]
 Let me try to reach 44 using these numbers.

First attempt:
99 - 48 = 51
51 - 46 = 5
5 * 11 = 55 (too high)

Second attempt:
48 - 46 = 2
99 - 11 = 88
88 / 2 = 44 (This works!)

Let me verify:
1. 48 - 46 = 2
2. 99 - 11 = 88
3. 88 / 2 = 44

This works! Let me write it formally.
Let me double-check the final answer with the calculator.
<tool>(99 - 11) / (48 - 46)</tool>

<obs>= 44</obs>
That equals 44, so the answer is correct.
</think>
<answer>(99 - 11) / (48 - 46)</answer>
\end{Verbatim}

\paragraph{Analysis.}
Plain RLOO repeatedly explores nearby subtractive patterns such as $99-48$ and $99-46$, but it does not identify the key decomposition $48-46=2$ and $99-11=88$. Tool-RLOO finds this decomposition and then uses the external tool at the verification point:
\begin{Verbatim}[fontsize=\footnotesize]
<tool>(99 - 11) / (48 - 46)</tool>
<obs>= 44</obs>
\end{Verbatim}
The tool is useful here because it converts a proposed plan into a checked arithmetic result before the final answer is emitted. This helps distinguish a valid expression from the many near-miss expressions generated in the plain RLOO trace.

\section{Implementation Details}
\label{sec:imp_details}
All RL experiments were run on a single H100 GPU. RLOO without tool use
  took 5.73 hours for 100 training steps, while Tool-RLOO with the matched 100-step training budget took 9.21 hours. The extended Tool-RLOO run continued from step 100 to step 200, requiring an additional 6.97 hours; therefore, the full 200-step Tool-RLOO training took 16.15 hours in total. Tool-RLOO++ took 5.29 hours for 100 steps, Tool-GRPO took about 5 hours, and Tool-DAPO took 6.29 hours for 100 steps. It should be noted that Tool-GRPO significantly deviates from SFT model after about 35 trianing steps, thus the Tool-GRPO is taken at the checkpoint with training step as 30.
  
  \begin{table}[htbp]
  \centering
  \small
  \caption{Hyperparameters for RLOO and Tool-RLOO. The two settings share the same
  RL hyperparameters; Tool-RLOO additionally enables tool execution during
  rollout.}
  \label{tab:hparams_rloo}
  \begin{tabular}{lc}
  \toprule
  Hyperparameter & Value \\
  \midrule
  Initial policy & SFT checkpoint \\
  Dataset & Countdown / Tool-Countdown \\
  Tool execution & No / Yes \\
  Max turns & -- / 10 \\
  Training steps & 100/200 \\
  Batch size & 128 \\
  Group size & 8 \\
  Gradient accumulation steps & 128 \\
  Learning rate & $1\times10^{-5}$ \\
  LR schedule & constant \\
  Warmup ratio & 0.0 \\
  Weight decay & $1\times10^{-4}$ \\
  Gradient clipping & 0.0 \\
  Entropy coefficient & $1\times10^{-3}$ \\
  KL coefficient & $1\times10^{-3}$ \\
  Temperature & 1.0 \\
  Top-$p$ & 1.0 \\
  Top-$k$ & -1 \\
  Min-$p$ & 0.0 \\
  Max generation tokens & 1024 \\
  Max model length & 2048 \\
  Checkpoint interval & 10 steps \\
  Policy-ratio clipping & No PPO-style clipping \\
  \bottomrule
  \end{tabular}
  \end{table}

  \begin{table}[htbp]
  \centering
  \small
  \caption{Hyperparameters for Tool-RLOO++. Tool-RLOO++ follows Tool-RLOO and
  modifies the advantage normalization.}
  \label{tab:hparams_rloo_pp}
  \begin{tabular}{lc}
  \toprule
  Hyperparameter & Value \\
  \midrule
  Initial policy & Tool-SFT checkpoint \\
  Dataset & Tool-Countdown \\
  Tool execution & Yes \\
  Max turns & 10 \\
  Training steps & 100 \\
  Batch size & 128 \\
  Group size & 8 \\
  Gradient accumulation steps & 128 \\
  Learning rate & $1\times10^{-5}$ \\
  LR schedule & constant \\
  Warmup ratio & 0.0 \\
  Weight decay & $1\times10^{-4}$ \\
  Gradient clipping & 0.0 \\
  Entropy coefficient & $1\times10^{-3}$ \\
  KL coefficient & $1\times10^{-3}$ \\
  Temperature & 1.0 \\
  Top-$p$ & 1.0 \\
  Top-$k$ & -1 \\
  Min-$p$ & 0.0 \\
  Max generation tokens & 1024 \\
  Max model length & 2048 \\
  Advantage normalization & Batch normalization \\
  Advantage target std. & 0.3 \\
  Policy-ratio clipping & No PPO-style clipping \\
  Checkpoint interval & 10 steps \\
  \bottomrule
  \end{tabular}
  \end{table}
  
\begin{table}[htbp]
    \centering
    \small
    \caption{Hyperparameters for Tool-GRPO. The trainer is a faithful single-update
    GRPO (DeepSeekMath, $\mu=1$): the importance ratio is identically 1 and PPO-style
    clipping is inert.}
    \label{tab:hparams_grpo}
    \begin{tabular}{lc}
    \toprule
    Hyperparameter & Value \\
    \midrule
    Initial policy & SFT checkpoint \\
    KL reference & Frozen SFT checkpoint \\
    Dataset & Tool-Countdown \\
    Tool execution & Yes \\
    Max tool calls & 5 \\
    Training steps & 250 \\
    Batch size & 128 \\
    Group size & 8 \\
    Gradient accumulation steps & 128 \\
    Inner epochs ($\mu$) & 1 \\
    Learning rate & $1\times10^{-5}$ \\
    LR schedule & constant \\
    Warmup ratio & 0.0 \\
    Weight decay & $1\times10^{-4}$ \\
    Gradient clipping & 1.0 \\
    Entropy coefficient & $1\times10^{-3}$ \\
    KL coefficient ($\beta$) & 0.01 \\
    Advantage & group-relative $(r-\mu)/(\sigma+10^{-4})$ \\
    Temperature & 0.6 \\
    Top-$p$ & 0.95 \\
    Top-$k$ & 20 \\
    Min-$p$ & 0.0 \\
    Max generation tokens & 1024 \\
    Max prompt length & 512 \\
    Max model length & 4096 \\
    Checkpoint interval & 10 steps \\
    Policy-ratio clipping & $\epsilon=0.2$ (inert at $\mu=1$) \\
    \bottomrule
    \end{tabular}
    \end{table}
    
  \begin{table}[htbp]
  \centering
  \small
  \caption{Hyperparameters for Tool-DAPO.}
  \label{tab:hparams_dapo}
  \begin{tabular}{lc}
  \toprule
  Hyperparameter & Value \\
  \midrule
  Initial policy & Tool-SFT checkpoint \\
  Dataset & Tool-Countdown \\
  Tool execution & Yes \\
  Max turns & 10 \\
  Training steps & 100 \\
  Batch size & 128 \\
  Group size & 8 \\
  Gradient accumulation steps & 128 \\
  Learning rate & $2\times10^{-5}$ \\
  LR schedule & constant \\
  Warmup ratio & 0.0 \\
  Weight decay & $1\times10^{-4}$ \\
  Gradient clipping & 1.0 \\
  Entropy coefficient & $1\times10^{-3}$ \\
  KL coefficient & $1\times10^{-3}$ \\
  Temperature & 1.0 \\
  Top-$p$ & 1.0 \\
  Top-$k$ & -1 \\
  Min-$p$ & 0.0 \\
  Max generation tokens & 1024 \\
  Max model length & 3072 \\
  Advantage normalization & Group normalization \\
  Policy-ratio clipping & $[1-0.2,\,1+0.28]$ \\
  Dynamic sampling & Yes \\
  Max resampling iterations & 8 \\
  Overlong penalty & Disabled \\
  Checkpoint interval & 10 steps \\
  \bottomrule
  \end{tabular}
  \end{table}
\end{document}